\pdfoutput=1

\documentclass[11pt]{article}

\usepackage{acl}

\usepackage{times}
\usepackage{latexsym}
\usepackage{graphicx}
\usepackage{pifont}
\usepackage{booktabs}
\usepackage{tabularx}
\usepackage{multirow}
\usepackage{enumitem}
\usepackage[T1]{fontenc}

\usepackage[utf8]{inputenc}

\usepackage{microtype}

\usepackage{inconsolata}
\newcommand{\instance}[1]{\textcolor{blue}{\fontfamily{pcr}\selectfont \textbf{#1}}}
\newcommand{\attribute}[1]{\textcolor{orange}{\fontfamily{pcr}\selectfont \textit{\textbf{#1}}}}
\newcommand{\constant}[1]{\textcolor{teal}{\fontfamily{pcr}\selectfont \textbf{#1}}}
\newcommand{\relation}[1]{\textcolor{purple}{\fontfamily{pcr}\selectfont \textit{\textbf{#1}}}}
\newcommand{\metric}[1]{\textit{#1}}

\title{\textsc{Rematch}: Robust and Efficient Matching of Local Knowledge Graphs \\to Improve Structural and Semantic Similarity}

\author{
Zoher Kachwala,\hspace{0.5em} Jisun An,\hspace{0.5em} Haewoon Kwak,\hspace{0.5em} Filippo Menczer\\
Observatory on Social Media\\
Indiana University\\
\texttt{zkachwal@iu.edu, \{jisun.an, haewoon\}@acm.org}
}

\begin{document}
\maketitle

\begin{abstract}
Knowledge graphs play a pivotal role in various applications, such as question-answering and fact-checking. Abstract Meaning Representation (AMR) represents text as knowledge graphs. Evaluating the quality of these graphs involves matching them structurally to each other and semantically to the source text. Existing AMR metrics are inefficient and struggle to capture semantic similarity. We also lack a systematic evaluation benchmark for assessing structural similarity between AMR graphs. To overcome these limitations, we introduce a novel AMR similarity metric, \metric{rematch}, alongside a new evaluation for structural similarity called RARE. Among state-of-the-art metrics, \metric{rematch} ranks second in structural similarity; and first in semantic similarity by 1--5 percentage points on the STS-B and SICK-R benchmarks. \metric{Rematch} is also five times faster than the next most efficient metric.\footnote{Our code for \metric{rematch} and RARE is publicly available at: \url{https://github.com/osome-iu/Rematch-RARE}}
\end{abstract}

\section{Introduction}

Knowledge graphs provide a powerful framework for multi-hop reasoning tasks, such as question answering and fact-checking \cite{yasunaga_qa-gnn_2021, vedula_face-keg_2021}. Even for closed-domain tasks like long-form question answering and multi-document summarization, knowledge graphs derived from individual documents --- referred to as \textit{local knowledge graphs} --- exhibit superior performance compared to plain text \cite{fan_using_2019}. This highlights the significance of automatically parsed knowledge graphs in both large-scale and fine-grained structured reasoning applications.

\begin{figure}[t]
    \centering
    \includegraphics[width=\columnwidth]{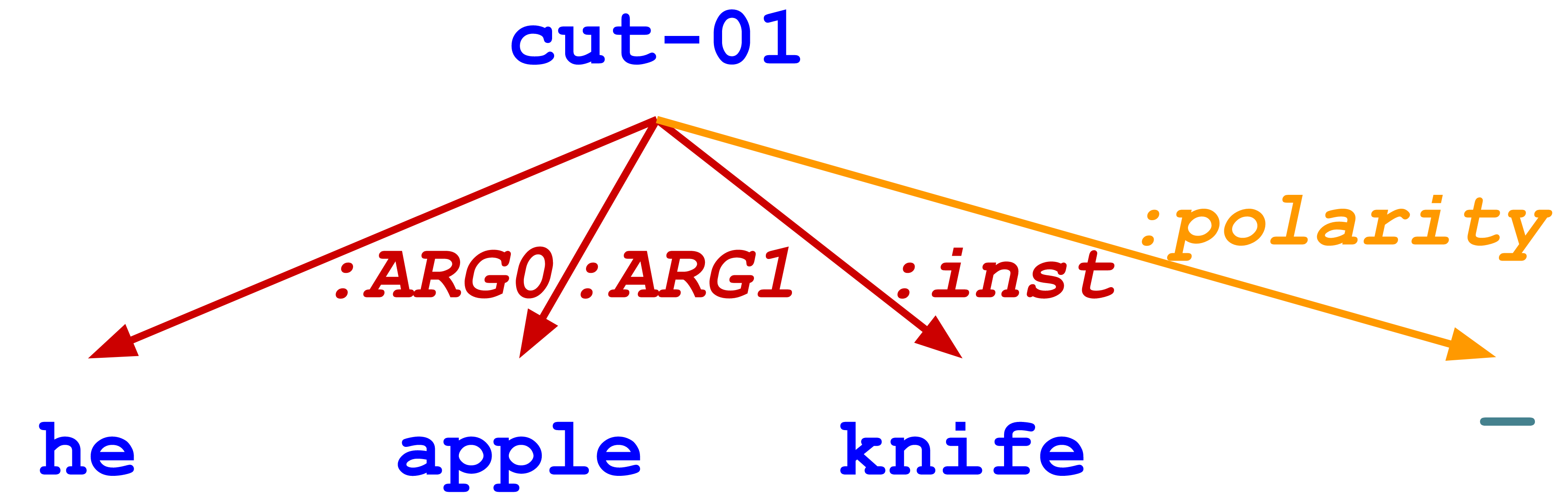}
    \caption{AMR for the sentence: \textit{``He did not cut the apple with a knife.''} Colors indicate AMR components: instances (blue), relations (red), constants (teal), and attributes (orange). The instance \instance{cut-01} is a verb frame that uses \relation{ARG0}, \relation{ARG1} and \relation{inst} to express the verb's agent (\instance{he}), patient (\instance{apple}), and instrument (\instance{knife}), respectively. The attribute \attribute{polarity} expresses the negation of the verb through the constant \constant{-}.}
    \label{fig:apple}
\end{figure}

The Abstract Meaning Representation (AMR) framework leverages acyclic, directed, labeled graphs to represent semantic meaning (knowledge) extracted from text \cite{banarescu_abstract_2013}. As illustrated in the example of Fig.~\ref{fig:apple}, AMRs capture the relationships between concepts and their roles in a sentence. They have been applied to a variety of natural language processing tasks, including summarization and question answering \cite{liu_toward_2015, hardy_guided_2018, bonial_dialogue-amr_2020, mitra_addressing_2016}. Recent work has also shown that AMRs can reduce hallucinations and improve performance in factual summarization and text classification tasks \cite{ribeiro_factgraph_2022, shou-etal-2022-amr}.



However, evaluating the quality of knowledge graphs like AMRs hinges critically on the ability to accurately measure similarity. This assessment must consider a dual perspective. Firstly, the similarity between two AMRs should reflect structural consistency, guaranteeing that the similarity between two AMRs aligns with the similarity of their structural connections. Secondly, AMRs should exhibit semantic consistency, ensuring that the similarity between two AMRs aligns with the similarity of the texts from which they are derived. Therefore, an effective AMR similarity metric must successfully account for both structural and semantic similarity, all while overcoming the resource-intensive nature of matching labeled graphs.


Current AMR similarity metrics fall short in several key areas. Firstly, their computational efficiency hinders the comparison of large AMRs extracted from documents \cite{naseem-etal-2022-docamr}. Secondly, these metrics struggle to accurately capture the semantic similarity of the underlying text from which AMRs are derived \cite{leung_semantic_2022}. Additionally, while recent efforts like BAMBOO \cite{opitz_weisfeiler-leman_2021} have evaluated metrics on AMR transformations, we still lack a large-scale benchmark to systematically evaluate the ability of AMR metrics to capture structural similarity.

Our work introduces a structural AMR benchmark called \textit{Randomized AMRs with Rewired Edges} (RARE) and proposes \metric{rematch}, a novel and efficient AMR similarity metric that captures both structural and semantic similarity. Compared to the state of the art, \metric{rematch} trails the best similarity metric on RARE by 1 percentage point and ranks first on the STS-B \cite{agirre_semeval-2016_2016} and SICK-R \cite{marelli_sick_2014} benchmarks by 1--5 percentage points. Additionally, \metric{rematch} is five times faster than the next most efficient metric.

\section{Background}

\subsection{Abstract Meaning Representations}

Abstract Meaning Representation (AMR) is a structural, explicit language model that utilizes directed, labeled graphs to capture the semantics of text \cite{banarescu_abstract_2013}. AMR is designed to be independent of surface syntax, ensuring that sentences with equivalent meanings are represented by the same graph.
An AMR comprises three fundamental components: instances, attributes, and relations.

\begin{enumerate}[wide, labelindent=0pt]

\item \textbf{Instances} are the core semantic concepts. Structurally, they are represented by nodes in the graph. AMRs have two types of instances. One utilizes \textit{PropBank} \cite{palmer_proposition_2005}, a dictionary of frames that map verbs and adjectives. The other comprises entities. Considering the sentence in Fig.~\ref{fig:apple}, \textit{``He did not cut the apple with a knife,''} the AMR contains a \textit{PropBank} instance \instance{cut-01} and three entity instances: \instance{he}, \instance{apple} and \instance{knife}. 

\item \textbf{Attributes} capture details about instances, such as names, numbers, and dates. These values are represented as constant nodes. Structurally, an attribute is identified in the graph as the edge from an instance node to a constant node. For example, in Fig.~\ref{fig:apple}, the attribute \attribute{polarity} is specified for the instance \instance{cut-01}, where \constant{-} is the constant that represents the negation of the verb.

\item \textbf{Relations} represent the connections between instances. In Fig.~\ref{fig:apple}, the instance \instance{cut-01} has three outgoing relations: \relation{ARG0}, \relation{ARG1}, and \relation{inst}. These come from \textit{PropBank}'s \instance{cut-01} frame and link to the agent (\instance{he}), the patient (\instance{apple}), and the instrument (\instance{knife}), respectively.

\end{enumerate}

\subsection{AMR Similarity}

Graph isomorphism is a test to determine whether two graphs are structurally equivalent. 
The class-wise isomorphism testing with limited backtracking (CISC) algorithm efficiently identifies isomorphic relationships in labeled graphs \cite{hsieh_efficient_2006}, such as AMRs. But a pair of AMRs may not have the same number of nodes, which violates a key assumption of graph isomorphism. A more appropriate approach is subgraph isomorphism, which determines whether a smaller graph is isomorphic to a subgraph of a larger graph. Subgraphs of directed acyclic graphs, like AMRs, can be enumerated in polynomial time \cite{peng_enumerating_2018}, enabling efficient application of the CISC test to each pair of smaller AMR and larger AMR subgraphs.
However, even if two AMRs are not subgraph-isomorphic, they may still exhibit similarities in meaning and structure. 
Next, we describe various existing approaches to measure the similarity between AMR graphs.

\subsection{AMR Similarity Metrics}

\begin{figure*}[t]
    \centering
    \includegraphics[width=\textwidth]{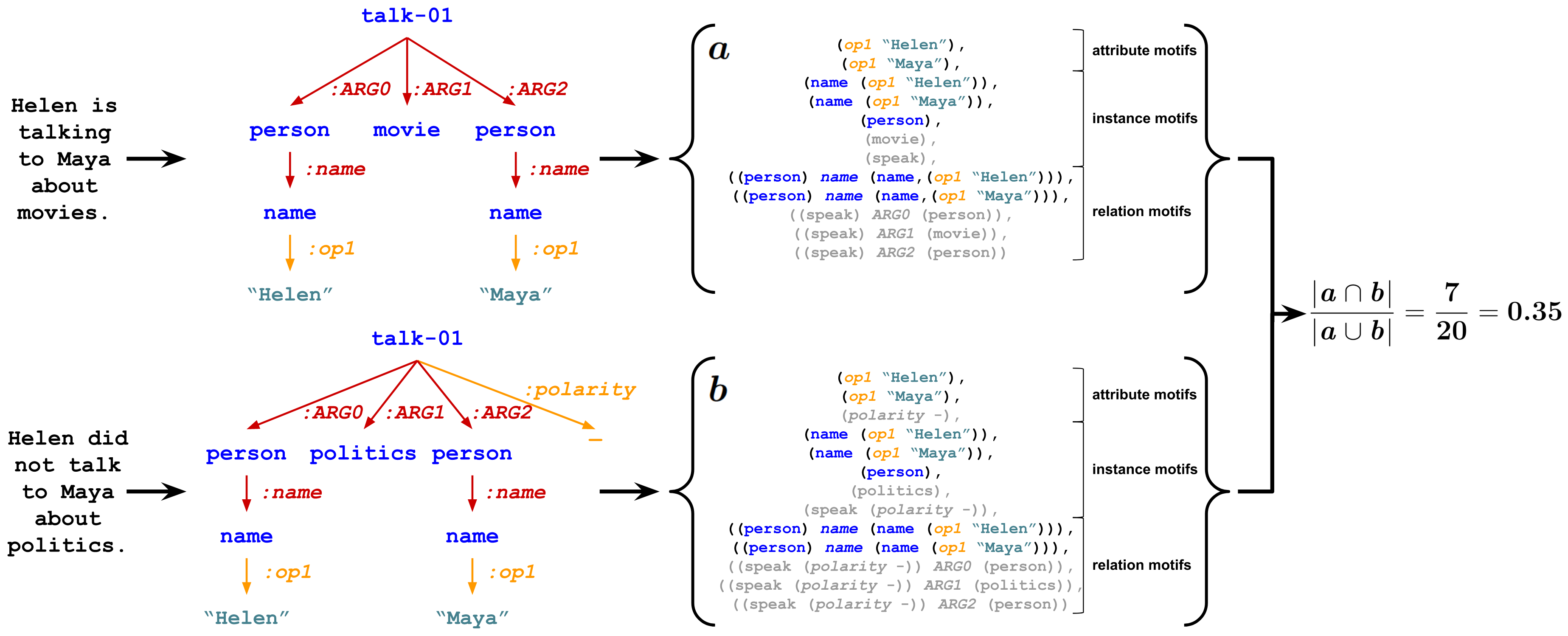}
    \caption{An example of \metric{rematch}'s similarity calculation for a pair of AMRs. After AMRs are parsed from sentences, \metric{rematch} has a two-step process to calculate similarity. First, sets of motifs are generated. Second, the two sets are used to calculate the Jaccard similarity (intersecting motifs shown in color).}
    \label{fig:rematchflow}
\end{figure*}

\subsubsection{Smatch}

\metric{Smatch} is a prominent tool for evaluating AMR parsers \cite{cai-knight-2013-smatch}. It establishes AMR alignment by generating a one-to-one node mapping, considering node and edge labels. To efficiently explore this vast mapping space, \metric{smatch} employs a hill-climbing heuristic.

\subsubsection{S2match}

Similar to \metric{smatch}, \metric{s2match} \cite{opitz_amr_2020} also establishes a node alignment between two AMRs. However, instead of relying on AMR labels, \metric{s2match} utilizes GloVe word embeddings \cite{pennington_glove_2014}. To address the extensive search space, it uses the same hill-climbing heuristic adopted by \metric{smatch}.

\subsubsection{Sembleu}

\metric{Sembleu} generates path-based n-grams from AMRs by leveraging node and edge labels \cite{song_sembleu_2019}. The final similarity score for an AMR pair is determined by calculating the BLEU score \cite{papineni_bleu_2002} between their n-grams. By avoiding a one-to-one node alignment, \metric{Sembleu} efficiently bypasses the issue of exploring a large search space.

\subsubsection{WLK}

The Weisfeiler-Leman Kernel (\metric{WLK}) and Wasserstein Weisfeiler-Leman Kernel (\metric{WWLK}) for AMRs also utilize graph features for computing similarity \cite{opitz_weisfeiler-leman_2021}. \metric{WLK} first constructs node features by recursively aggregating AMR node and edge labels. Then it generates a frequency-based feature vector for each AMR and calculates a similarity score using their inner product. \metric{WWLK} extends \metric{WLK} with features based on aggregated node embeddings (GloVE) instead of node labels. Since WWLK is a supervised metric, we do not consider it in our evaluation.

\section{Methods}

In this work, we propose \metric{rematch}, an AMR similarity metric that aims to capture both the structural and semantic overlap between two AMRs. 


A straightforward approach to match two labeled graphs involves identifying the alignment between node labels. However, labeled graphs often contain duplicate labels, necessitating an exhaustive exploration of all one-to-one combinations among nodes within the same label group to determine the optimal match. The resulting matching complexity hinges on the size of node groups with shared labels. 
This is why algorithms like \metric{smatch} and \metric{s2match} do not scale well to large AMRs, where these node groups can be large.

Graph features constructed using an ordered concatenation of edge-node bi-grams are utilized in both isomorphism tests like the CISC and kernels like Weisfeiler-Leman \cite{shervashidze_weisfeiler-lehman_2011}. This approach is effective: it consistently produces smaller node groups compared to those based solely on node labels. Matching between two graphs is significantly accelerated as a result.

Inspired by this idea of exploiting graph features for efficiency, \metric{rematch} computes the similarity between two AMRs by analyzing the overlap of semantically rich features, which we call \textit{motifs}. 
Unlike the ordered graph partitions used by CISC and Weisfeiler-Leman Kernel, which rely on node and edge labels, AMR \textit{motifs} are unordered graph partitions that leverage AMR instances, attributes, and relations. 
This approach allows \metric{rematch} to capture meaning across three semantic levels: specific facts (attributes), main concepts (instances), and the relationships among concepts (relations). 
Fig.~\ref{fig:rematchflow} illustrates \metric{rematch} through an example. 
Next, we delve into the three orders of semantic motifs that we use for \metric{rematch}. We extract these motifs using the Python package Penman \cite{goodman_penman_2020}. 

\begin{enumerate}[wide, labelindent=0pt]

\item \textbf{Attribute motifs} are pairs of attributes and constants associated with AMR instance nodes. For the bottom AMR in Fig.~\ref{fig:rematchflow}, \instance{talk-01} has attribute motif (\attribute{polarity}~\constant{-}), indicating a negation. The first \instance{name} has the attribute motif (\attribute{op1}~\constant{"Helen"}) and the second \instance{name} has (\attribute{op1}~\constant{"Maya"}), identifying the name values. The remaining instances do not have any attributes.

\item \textbf{Instance motifs} leverage \textit{Verbatlas}, a resource that maps \textit{PropBank} frames to more generalized frames \cite{di_fabio_verbatlas_2019}. If an instance in the AMR corresponds to a Verbatlas frame, the latter is used instead. Otherwise, the original \textit{PropBank} instance is retained. For example, in Fig.~\ref{fig:rematchflow}, \instance{talk-01} is replaced by the more generalized Verbatlas frame \instance{speak}. 
The generation of instance motifs follows two approaches. If an instance lacks associated attributes, the instance itself serves as its motif. However, if attributes are present, instance motifs are constructed by combining the instance with each of its attribute motifs. For the bottom AMR in Fig.~\ref{fig:rematchflow}, the instance motif for \instance{talk-01} is (\instance{speak}~(\attribute{polarity}~\constant{-})), indicating a negation of the verb. For the two \instance{person} instances and the \instance{politics} instance, the instances themselves become their motifs, namely (\instance{person}) and (\instance{politics}). Finally, the instance motifs for the two \instance{name} instances are (\instance{name}~(\attribute{op1}~\constant{"Helen"})) and (\instance{name}~(\attribute{op1}~\constant{"Maya"})) respectively, identifying the names in the conversation.

\item \textbf{Relation motifs} are constructed for relation edges in an AMR graph. Each relation motif comprises three elements: an instance motif of the source instance, the relation label, and an instance motif of the target instance. A relation can have multiple relation motifs, one for each unique combination of source and target instance motifs. 
For the bottom AMR in Fig.~\ref{fig:rematchflow}, the relation motifs for 
\relation{ARG0}, \relation{ARG1} and \relation{ARG2} are:  
((\instance{speak}~(\attribute{polarity}~\constant{-}))~\relation{ARG0}~\instance{person}), indicating a person is the speaker of the conversation; ((\instance{speak}~(\attribute{polarity}~\constant{-}))~\relation{ARG1}~\instance{politics}), indicating that the topic of conversation is politics; and ((\instance{speak}~(\attribute{polarity}~\constant{-}))~\relation{ARG2}~\instance{person}), indicating that a person is the recipient of the conversation. 
For the two \relation{name} relations, the motifs are: ((\instance{person})~\relation{name}~(\attribute{op1}~\constant{"Helen"}))), identifying "Helen" as the name of one person; and ((\instance{person})~\relation{name}~(\attribute{op1}~\constant{"Maya"})), identifying "Maya" as the name of the other person.

\end{enumerate}

Each AMR is represented by the union of its instance, relation, and attribute motifs. The \metric{rematch} score between two AMRs is determined by calculating the Jaccard similarity between their respective motif sets, as illustrated in Fig.~\ref{fig:rematchflow}.

\section{Evaluation}

We evaluate the effectiveness of \metric{rematch} on three types of similarity: structural similarity, semantic similarity, and BAMBOO \cite{opitz_weisfeiler-leman_2021}, a hybrid benchmark that modifies AMR semantics through structural transformations. Additionally, we assess the efficiency of \metric{rematch}. 

\subsection{Structural Similarity (RARE)}

Given that AMRs are graphical representations of text, an AMR similarity metric should be sensitive to structural variations between AMRs, even if its labels remain unchanged.

Since there is no established evaluation of AMR metrics on structural similarity, we have developed a new benchmark dataset called \textit{Randomized AMRs with Rewired Edges} (RARE). RARE consists of English AMR pairs with similarity scores that reflect the structural differences between them. 

In the construction of RARE, we adopt an iterative randomization technique commonly used for graph rewiring. This involves repeatedly selecting a random pair of directed edges and swapping either their source or target nodes to establish new connections. This way each node's in-degree and out-degree are preserved. In applying this approach to AMRs, we swap a random pair of edges between either attributes or relations. 
This allows us to quantify the structural changes made to the AMR through the number of swapped edges. 

RARE does not add or remove edges as these modifications would amount to adding or removing information. 
Systematic edge insertion or deletion would also introduce additional complications, such as having to decide the set of edges that could be added or removed while keeping the network connected. 
By swapping edges alone, we guarantee that the AMRs being compared have the same information in terms of size, density, and connectivity. 

We generate a spectrum of modified graphs from an original AMR, ranging from the unchanged graph to one where all edges are rewired, subject to some constraints that preserve the integrity of AMRs:

\begin{enumerate}[wide, labelindent=0pt]
    \item \textbf{Structural Constraints.} AMRs are acyclic, connected graphs that allow no multiedges (more than one edge between the same pair of nodes). To preserve these properties during the rewiring process, pairs of swapped edges must maintain these constraints in the modified AMR.
    
    \item \textbf{Semantic Constraints.} These constraints relate to swapping attributes and relations:
    
    \begin{enumerate}
        \item Attributes have an inherent connection with constants in AMRs. Hence, while rewiring a pair of attribute edges, only the source instance node should be swapped. This restriction ensures that the association between the attribute and its corresponding constant remains intact. For example, the constant node \constant{-} should remain associated solely with the attribute edge \attribute{polarity}.
        
        \item Relations in AMRs connect two instances. When rewiring a pair of relation edges, only the target instance node should be swapped. This restriction maintains the association between the relation's source instance and the relation itself. For example, \textit{PropBank} instances have a predefined set of relations with which they can be associated. The instance node \instance{talk-01} can only be associated with edges \relation{ARG0}, \relation{ARG1}, and \relation{ARG2}.
    \end{enumerate}
\end{enumerate}

Each pair of AMRs, consisting of an original AMR $G$ with $E$ edges and its corresponding rewired AMR $G'$ with $E'$ swapped edges, is annotated with the following similarity score:
\begin{equation}
similarity(G, G') = \frac{|E| - |E'|}{|E|} \label{eq:similarity}
\end{equation}

To generate the RARE benchmark, we licensed the English AMR Annotation 3.0 \cite{knight_kevin_abstract_2020} containing 59,255 human-created AMRs. Using the process described above, we get 563,143 rewired AMR pairs annotated with similarity scores per Eq.~\ref{eq:similarity}. Since the original AMR Annotation 3.0 corpus has an unusual training-development-testing split, we merge, shuffle, and re-split AMR 3.0 into training (47,404), development (5,925), and test (5,926) sets to get an 80-10-10 split ratio that is more consistent with standard benchmarks. The resulting RARE training-development-test sizes are 450,067, 56,358, and 56,718, respectively. The creation of training and development splits could facilitate the future development of supervised AMR metrics. For the current evaluation, AMR structural similarity metrics are evaluated on the RARE test split. 

We evaluate a similarity metric by computing the Spearman correlation between its scores and the ground truth values from Eq.~\ref{eq:similarity}, across a set of pairs of original and modified AMRs. We refer to this as the \textit{structural consistency} of the metric.

\subsection{Semantic Similarity}

A fundamental tenet of AMRs is that if two pieces of text are semantically related, their corresponding AMRs should exhibit a degree of similarity. But a metric could deem two AMRs similar even when their textual sources have very different meanings. As an example, for two completely unrelated sentences ``Spanish bulls gore seven to death'' and ``Obama queries Turnbull over China port deal,'' \metric{smatch} assigns a non-zero score due to the similarity in their argument structure \cite{leung-etal-2022-semantic}. To tease out such shortcomings, we evaluate each AMR similarity metric by considering many pairs of sentences. For each pair, we compare the similarity generated by the metric for the corresponding AMRs to a ground-truth similarity score between the sentences generated by human annotations. 

We utilize two standard sentence similarity benchmarks for English: STS-B \cite{agirre_semeval-2016_2016} and SICK-R \cite{marelli_sick_2014}. To account for variations in AMR parsing accuracy, we employ four different AMR parsers: \textit{spring} \cite{bevilacqua_one_2021}, \textit{amrbart} \cite{bai_graph_2022}, \textit{structbart} \cite{drozdov_inducing_2022-1}, and the \textit{maximum Bayes \metric{smatch} ensemble} \cite{lee_maximum_2022}.

Given a set of sentence pairs and corresponding AMR pairs, we evaluate a similarity metric by computing the Spearman correlation between its scores for the AMR pairs and the human-annotated similarity values for the sentence pairs. We refer to this as the \textit{semantic consistency} of the metric.
Note that semantic consistency can be used to evaluate any similarity method for sentences, not only AMR-based ones. 
For both structural and semantic consistency, we use Spearman rather than Pearson correlation because we do not assume that the similarity values are normally distributed. 

\subsection{Hybrid Similarity (BAMBOO)}

In addition to the structural and semantic consistency discussed earlier, we evaluate the robustness of AMR metrics using the \textit{Benchmark for AMR Metrics Based on Overt Objectives}, or BAMBOO \cite{opitz_weisfeiler-leman_2021}. BAMBOO assesses the ability of AMR similarity metrics to capture semantic similarity between English sentences while modifying the structure of the corresponding AMRs.

BAMBOO incorporates three types of graph modifications: synonym replacement, reification, and role confusion. 
Consider the example sentence ``He lives in the attic,'' represented by an AMR where the node \instance{live-01} connects to nodes \instance{he} and \instance{attic} via the edges \relation{ARG0} and \relation{location}, respectively. 
Synonym replacement swaps \textit{PropBank} instances with equivalent terms. In the example, \instance{live-01} might be replaced by \instance{reside-01}. 
Reification transforms a relation into a new instance. In the example, the \relation{location} edge might be replaced by a new node \instance{be-located-at-91} connected to \instance{live-01} and \instance{attic} via new \relation{ARG1} and \relation{ARG2} edges, respectively. 
Finally, role confusion swaps relation roles. In the example, the relations \relation{location} and \relation{ARG0} might be swapped such that the modified AMR would represent the sentence ``The attic lives in him.'' 
BAMBOO applies these modifications to the original train, test and dev splits of the STS-B, SICK-R, and PARA \cite{dolan_automatically_2005} datasets.

Given a set of modified AMR pairs, BAMBOO evaluates an AMR metric by the Spearman correlation\footnote{The original formulation of BAMBOO \cite{opitz_weisfeiler-leman_2021} used Pearson correlation. Here we use Spearman because, as for structural and semantic consistency, we do not assume that the similarity values are normally distributed.} between its scores and the similarity between the corresponding sentence pairs. We call this \textit{hybrid consistency} of the metric.

\subsection{Efficiency}

As discussed earlier, the computational complexity associated with node alignment is a crucial challenge for comparing AMRs. To address this issue, we evaluate the search spaces explored by various metrics and the required runtime. 

We establish a realistic test bed using the AMR Annotation 3.0 once again. For this evaluation, we randomly sampled 500,000 pairs from the ${59,255}\choose{2}$ possible AMR combinations.
For each pair of AMRs $(G_1, G_2)$, the search spaces for node alignment algorithms like \metric{smatch} and \metric{s2match} is 
\begin{equation}
search(G_1, G_2) =  \prod_{n_i\in G_1}|M_{G_2}(n_i)|
\label{eq:node_map}
\end{equation}
where $M_{G_2}(n_i)$ denotes the set of matching candidates in $G_2$ for node $n_i$.
For feature-based algorithms, like \metric{sembleu}, \metric{WLK}, and \metric{rematch}, we record the search space using
\begin{equation}
search(G_1, G_2) =  |\mathcal{F}(G_1)| \cdot |\mathcal{F}(G_2)| 
\label{eq:feature}
\end{equation}
where $\mathcal{F}(G)$ denotes the feature set for graph $G$.
For each pair of AMRs, we also record the runtime.

\section{Results}

\subsection{Structural Consistency}

\begin{table}
\centering
\begin{tabular}{lc}
\toprule
AMR Metric & RARE\\
\hline
\metric{smatch}  & \textbf{96.57} \\
\metric{s2match}  & 94.11 \\
\metric{sembleu}  & 94.83 \\
\metric{WLK}  & 90.39 \\
\hline
\metric{rematch} & 95.32 \\
\bottomrule
\end{tabular}
\caption{\label{structural-similarity}
Structural consistency of different AMR similarity metrics on the RARE test split.
}
\end{table}

Table~\ref{structural-similarity} reports on the structural consistency of the AMR similarity metrics on the RARE test split. We can see that \metric{smatch} performs the best, followed closely by \metric{rematch}, \metric{sembleu} and \metric{s2match}. The subpar performance of \metric{WLK} can be attributed to their reliance on features using all of a node's neighbors. This approach results in changes to node features regardless of the number of modified neighbors, failing to capture the nuances of neighborhood changes. 

\subsection{Semantic Consistency}

\begin{table*}
\centering
\begin{tabular}{lcccc|cccc}
\toprule
 & \multicolumn{4}{c}{STS-B} & \multicolumn{4}{c}{SICK-R}  \\
\hline
& \textit{spring} & \textit{amrbart} & \textit{sbart} & \textit{mbse} & \textit{spring} & \textit{amrbart} & \textit{sbart} & \textit{mbse}\\
\hline
\metric{smatch} & 53.84 & 54.67 & 54.73 & 55.16 & 58.69 & 58.89 & 58.70 & 57.84\\
\metric{s2match} & 56.60 & 57.15 & 57.54 & 57.64 & 58.09 & 58.56 & 58.42 & 57.58\\
\metric{sembleu} & n/a & 58.62 & 58.17 & 58.95 & 60.15 & 60.61 & 59.62 & 59.57\\
\metric{WLK} & 63.18 & 64.60 & 64.33 & 65.37 & 63.09 & 63.33 & 63.07 & 62.59\\
\hline
\metric{rematch} & \textbf{64.93} & \textbf{65.88} & \textbf{65.06} & \textbf{66.52} & \textbf{67.03} & \textbf{67.72} & \textbf{67.10} & \textbf{67.34}\\
\bottomrule
\end{tabular}
\caption{\label{semantic-similarity}
Semantic consistency of AMR similarity metrics and AMR parsers on the test splits of STS-B and SICK-R datasets. Best results are highlighted in bold. \metric{Sembleu} fails to parse some of the AMRs generated by \textit{spring}.
}
\end{table*}

\begin{table}
\centering
\begin{tabular}{lcc}
\toprule
Similarity Methods & STS-B & SICK-R  \\
\hline
GloVe (avg.) & 58.02 & 53.76 \\
RoBERTa (first-last avg.) & 58.55 & 61.63\\
AMR (\metric{rematch}) & 66.52 & 67.72\\
SimCSE-RoBERTa & \textbf{80.22} & \textbf{68.56}\\
\bottomrule
\end{tabular}
\caption{\label{semantic-similarity-2}
Comparison of similarity methods (AMR and non-AMR) on semantic consistency for the test splits of STS-B and SICK-R datasets.
}
\end{table}

Table~\ref{semantic-similarity} reports on the semantic consistency of the similarity metrics for different AMR parsers. \metric{Rematch} outperforms all other metrics by 1--5 percentage points, across all parsers and benchmarks. 
The \textit{mbse} and \textit{amrbart} parsers perform best for the STS-B and SICK-R datasets, respectively. 

So far we have focused on methods that use AMRs to calculate the semantic similarity between sentences.  
Table~\ref{semantic-similarity-2} reports on the evaluation of alternative similarity methods on the same benchmarks. 
Like AMR-based methods, these are also unsupervised (not trained specifically) for textual semantic similarity. 
AMR outperforms some representations like GloVe and RoBERTa but lags behind the state-of-the-art method SimCSE \cite{gao_simcse_2022}.

\subsection{Hybrid Consistency}

\begin{table*}
\centering
\setlength{\tabcolsep}{2.75pt}
\begin{tabular}{lccc|ccc|ccc|ccc|c}
\toprule
& \multicolumn{3}{c}{Main} & \multicolumn{3}{c}{Reification} & \multicolumn{3}{c}{Synonym Replace} & \multicolumn{3}{c}{Role Confusion} & Avg.\\
\hline
& {\small STS-B} & {\small SICK-R} & {\small PARA} & {\small STS-B} & {\small SICK-R} & {\small PARA} & {\small STS-B} & {\small SICK-R} & {\small PARA} & {\small STS-B} & {\small SICK-R} & {\small PARA}\\
\hline
\metric{smatch} & 53.01 & 57.65 & 40.96 & 53.02 & 59.74 & 40.08 & 51.76 & 55.42 & 39.60 & \textbf{54.03} & \textbf{75.20} & 24.78 & 50.44\\
\metric{s2match} & 55.87 & 57.38 & \textbf{41.92} & 55.41 & 59.46 & \textbf{40.78} & 54.86 & 56.12 & \textbf{40.59} & 48.23 & 73.89 & \textbf{26.19} & \textbf{50.89}\\
\metric{sembleu} & 57.02 & 58.76 & 31.95 & 54.73 & 59.92 & 31.92 & 53.42 & 54.66 & 27.95 & 45.69 & 66.74 & 21.36 & 47.01\\
\metric{WLK} & 63.68 & 62.32 & 35.18 & 61.31 & \textbf{63.03} & 35.65 & 57.90 & 56.60 & 31.43 & 44.72 & 66.39 & 17.46 & 49.64\\
\hline
\metric{rematch} & \textbf{64.72} & \textbf{66.54} & 34.88 & \textbf{63.49} & 62.55 & 35.82 & \textbf{59.75} & \textbf{61.54} & 32.70 & 42.38 & 67.28 & 15.37 & 50.59\\
\bottomrule
\end{tabular}
\caption{\label{bamboo}
Hybrid consistency of AMR similarity metrics on the test split of the BAMBOO benchmark, for the three kinds of modifications, no modification (main) and the overall average. The best results are highlighted in bold.
}
\end{table*}

Table~\ref{bamboo} reports on the hybrid consistency of AMR similarity metrics on the four different tests of BAMBOO, across three different datasets. The results vary considerably across graph modifications and datasets; none of the methods is a clear winner. \metric{Rematch} achieves best results in three out of twelve tests and lags slightly behind \metric{s2match} on average.

\subsection{Efficiency}


\begin{figure*}[t]
    \centering
    \includegraphics[width=\textwidth]{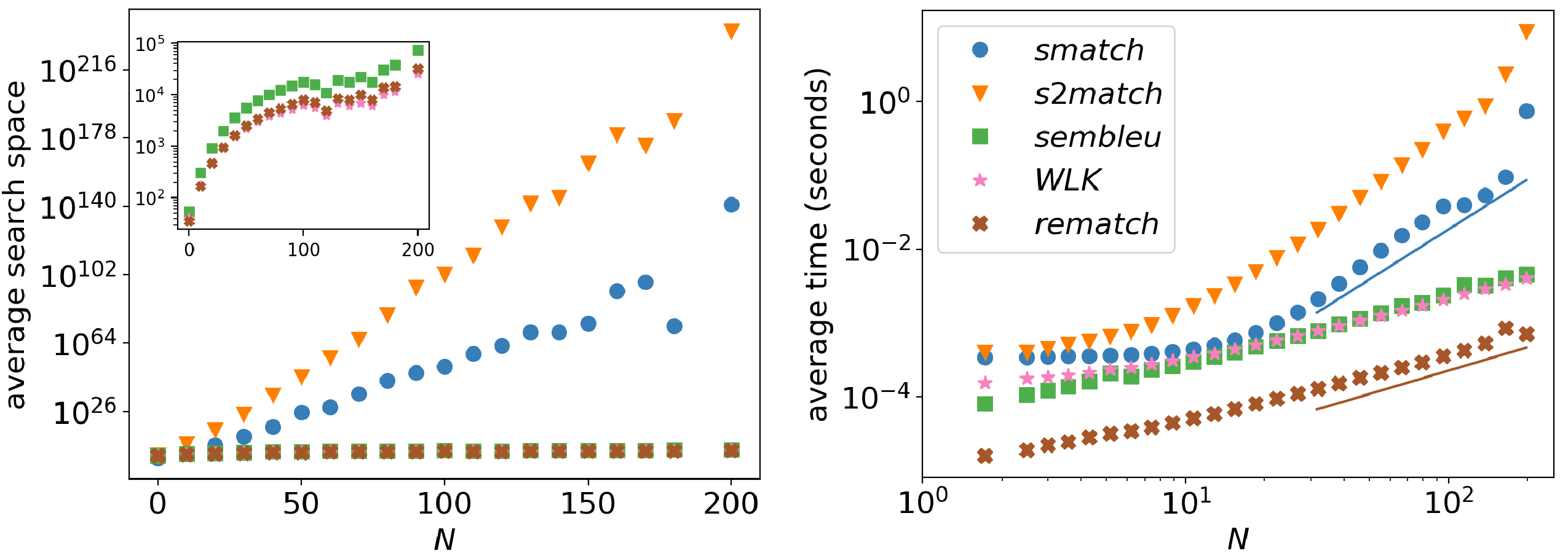}
    \caption{Average search space (left) and runtime (right) on a random sample of 500k pairs from  AMR Annotation 3.0. $N$ denotes the average size of each AMR pair. 
    The inset zooms in on \metric{sembleu}, \metric{WLK}, and \metric{rematch}, which cannot be distinguished in the log-linear plot. 
    The lines on the runtime plot indicate approximate fits for $N > 10^{1.5}$, which on the log-log scale represent polynomial time complexity. The slopes indicate that the runtime scales quadratically for \metric{smatch}  $O(N^{2.25})$ and linearly for \metric{rematch} $O(N)$.}
    \label{fig:time_space}
\end{figure*}

Fig.~\ref{fig:time_space} shows the search spaces explored by AMR metrics for increasing values of $N$, the average size of each pair of AMRs. The size of each AMR is determined by the sum of the number of instances, attributes, and relations. Approaches that find node alignment between AMRs, like \metric{smatch} and \metric{s2match}, explore search spaces that grow exponentially with $N$. Feature-based methods, like \metric{sembleu}, \metric{WLK}, and \metric{rematch}, in contrast, explore significantly smaller spaces.

Fig.~\ref{fig:time_space} also shows the runtimes for increasing $N$. By using a hill-climbing heuristic, node-alignment metrics effectively overcome the exponentially growing search spaces. However, they are significantly less efficient compared to feature-based metrics. For large values of $N$, \metric{smatch} and \metric{s2match} display an approximately quadratic time complexity. \metric{Sembleu}, \metric{WLK}, and \metric{rematch}, on the other hand, demonstrate a linear complexity. 

In terms of absolute runtime on the test bed, \metric{rematch} is the fastest metric, with a runtime of 51 seconds. This is five times faster than \metric{sembleu}, which took 275 seconds. \metric{Smatch}, \metric{s2match}, and \metric{WLK} trailed further behind, requiring 927, 7718, and 315 seconds. All metrics executed the test bed on a single 2.25~GHz core. \metric{Rematch}, \metric{sembleu}, and \metric{smatch} needed 0.2~GB of RAM, whereas \metric{s2match} and \metric{WLK} required 2~GB and 30~GB, respectively. We leave the efficiency comparison against non-AMR similarity methods like GloVe, RoBERTa and SimCSE as future work.

\subsection{Ablation Study}

\begin{table}[t]
\centering
\setlength{\tabcolsep}{3.5pt}
\begin{tabular}{lrrr}
\toprule
& \multicolumn{1}{c}{RARE} & \multicolumn{1}{c}{STS-B} & \multicolumn{1}{c}{SICK-R} \\
\hline
\metric{rematch} & 95.01 & 73.95 & 71.01 \\
\hline
 $-$ attribute & $-$00.85 & $-$00.40 & $-$00.09\\
 $-$ instance & $+$00.08 & $-$06.34 & $-$07.12\\
 $-$ relation & $-$62.30 & $+$01.15 & $-$01.41\\
 $-$ attribute, instance & $+$01.18 & $-$16.78 & $-$07.32\\
 $-$ attribute, relation & $-$62.55 & $+$00.93 & $-$01.92\\
 $-$ instance, relation & $-$95.87 & $-$37.90 & $-$62.32\\
\hline
 labels & $-$72.89 & $-$08.08 & $-$07.30\\
\bottomrule
\end{tabular}
\caption{\label{ablation}
Ablation study of different motifs on structural (RARE) and semantic (STS-B, SICK-R) consistency. Dev splits of RARE and STS-B, and the trial split of SICK-R were used. The \textit{mbse} and \textit{amrbart} parsers were used for STS-B and SICK-R, respectively.}
\end{table}


To assess the impact of the three types of \metric{rematch} motifs --- attribute, instance, and relation --- on structural and semantic similarity, let us conduct an ablation study, in which we remove one or more types of motifs at a time. The results are presented in Table~\ref{ablation}. Instance motifs have the most significant influence on semantic similarity, particularly when combined with relation motifs. Conversely, relation motifs exert the strongest influence on structural similarity, especially when complemented by instance motifs.

To evaluate the overall effectiveness of motifs, we also assess the performance of \metric{rematch} through the use of AMR labels alone. For the bottom AMR in Fig.~\ref{fig:rematchflow}, the label set is \{\instance{talk-01}, \instance{person}, \instance{politics}, \instance{name}, \relation{ARG0}, \relation{ARG1}, \relation{ARG2}, \relation{name}, \constant{-}, \constant{"Helen"}, \constant{"Maya"},  \attribute{polarity},  \attribute{op1}\}. Note that \instance{person}, \instance{name}, and \attribute{op1} appear only once in the set. Similar to \metric{rematch} motifs, we calculate the Jaccard similarity between two AMR label sets.
As shown in Table~\ref{ablation}, the decline in structural consistency when using AMR labels is substantial, given the absence of structural information in the label sets. 
In contrast, the decline in semantic consistency is relatively modest, indicating that AMR labels play a significant role in capturing semantics.

\subsection{Error Analysis}


On structural consistency, we find that \metric{rematch} underperforms when RARE swaps attribute edges connected to instance nodes with many relations. While the change might seem minor (a single swapped edge), the nested motif structure of \metric{rematch} amplifies the difference: mismatches in attribute motifs extend to instance motifs and all connected relation motifs, leading to a significant discrepancy in the overall similarity score.

The nested nature of \metric{rematch} can also sometimes underestimate semantic similarity. For instance, consider the sentences ``Work into it slowly'' and ``You work on it slowly.'' The first sentence's AMR associates an \attribute{imperative} attribute with the verb \instance{work-01}. This feature is missing in the second sentence. Consequently, \metric{rematch} generates different instance and relation motifs, resulting in a lower similarity score compared to the ground-truth similarity.

More often, the nested motif generation grants \metric{rematch} an advantage in semantic consistency tasks: it allows \metric{rematch} to handle negation more effectively compared to other metrics. For example, the sentences ``You should do it'' and ``You should never do it'' have a lower similarity score in \metric{rematch} due to the presence of the negative (\constant{-}) \attribute{polarity} attribute.

\section{Conclusion}

This paper introduces \metric{rematch}, a novel and efficient metric for AMR similarity. \metric{Rematch} leverages semantic AMR motifs to outperform existing metrics in both semantic consistency and computational efficiency. Additionally, we present RARE, a new benchmark designed to evaluate the structural consistency of AMR metrics. Using RARE, we demonstrate the strong sensitivity of \metric{rematch} to structural changes in AMRs.

AMR matching was originally introduced to evaluate and enhance AMR parsers. 
Through improved matching, metrics like \metric{rematch} improve parsing, which \textit{indirectly} benefits downstream uses of AMRs. 
But \metric{rematch} shows that AMRs encode richer semantics than previously assumed. Thus, improved AMR matching also \textit{directly} benefits downstream applications, like semantic textual similarity. 

Future research should explore the full potential of AMRs for natural language understanding. Natural Language Inference (NLI) is a prime example, where AMR-based systems have already shown promise \cite{opitz_amr4nli_2023}. An even more intriguing direction would be to develop methods that perform NLI solely through AMR matching, capitalizing on the rich structure and semantics encoded within AMRs. 

\section{Limitations}

Current AMR metrics, including \metric{rematch}, have limitations for downstream tasks like semantic textual similarity. One key issue is their inability to capture similarity between words. This can lead metrics like \metric{rematch} to misclassify two sentences with different wordings but equivalent meaning as dissimilar. 
\metric{S2match} attempts to address this limitation by using word embeddings for node alignment, but our analysis shows that this approach offers minimal improvement in semantic consistency at a high computational cost.
Recently, this limitation was addressed by a novel self-supervised metric called AMRSim \cite{shou-lin-2023-evaluate}. It trains Siamese BERT models on flattened silver AMR pairs generated from one million sentences sampled from Wikipedia.

Another limitation of \metric{rematch} is that it uses motifs associated with single edges (paths of length one). While this approach works well for short-text semantic similarity, it might not capture the more complex semantics present in AMRs derived from longer documents. In other words, \metric{rematch} might struggle to compare the meaning of longer texts.

\paragraph{Acknowledgements.} 
We thank Ramón Fernandez Astudillo for helpful discussions. 
This work was supported in part by the NSF through an NRT Fellowship (grant 1735095), the 
Knight Foundation, and Craig Newmark Philanthropies. We also acknowledge the Lilly Endowment for computational support through the Indiana University Pervasive Technology Institute.

\bibliography{references, anthology}




\end{document}